# IRIS BIOMETRIC SYSTEM USING A HYBRID APPROACH

Abhimanyu Sarin

B.Tech EEE (Student), BITS Pilani, Dubai Campus, Dubai, UAE
absarin1004@gmail.com

## ABSTRACT

*Iris Recognition Systems are ocular- based biometric devices used primarily for security reasons. The complexity and the randomness of the Iris, amongst various other factors, ensure that this biometric system is inarguably an exact and reliable method of identification. The algorithm is responsible for automatic localization and segmentation of boundaries using circular Hough Transform, noise reductions, image enhancement and feature extraction across numerous distinct images present in the database. This paper delves into the various kinds of techniques required to approximate the pupillary and limbic boundaries of the enrolled iris image, captured using a suitable image acquisition device and perform feature extraction on the normalized iris image with the help of Haar Wavelets to encode the input data into a binary string format. These techniques were validated using images from the CASIA database, and various other procedures were also tried and tested.*

## 1. INTRODUCTION

Today, we live in an age where our identity is what defines us more than anything else, but it is also a lot easier to get lost in the midst of the 7 billion people around us. This gives rise to more acute issues- mostly to do with counterfeiting and imitating another's self, a major security predicament .This is where biometric based recognition systems help us in ensuring the safety and security of the things that matter. The identity of a person is demarcated and stored by using an algorithm designed in a way to match the enrolled image(s) when an individual wants to log into the system again.

The most basic concept in any biometric system revolves around the basic processes of acquiring a high-resolution and feature-rich image, followed by detailed analysis of the desired part using image processing techniques and finally matching these details to a given input image. Iris Recognition systems use a very similar methodology.

First developed by Dr. John Daugman in the 1990s[1], who borrowed the idea from Flom and Safir's patented theoretical design, it has been greatly researched on since to make the automated system more efficient and versatile. Some of the main advantages of the this system is the organ itself- the Iris, a doughnut shaped colored structure in the eye is consistent of various features which is inarguably as unique as a fingerprint, if not more - a very rich, random interwoven texture called the "trabecular meshwork"[2]. These features are not chronologically perturbed and are genetically incoherent, meaning even twins have different eyes, and are also stable for a lifetime. It is also convenient in a sense that the eye is usually always there with a person, plus since it is an internal muscle, it relaxes when one dies, so even if it is removed it cannot be used fraudulently. Apart from being convenient, clean and secure, it is also very safe by being unobtrusive, as only an image of the eye is taken, not damaging any neural sensors in the eye.

Iris Biometric Identification systems have found major applications around the globe, it is being used in offices as an entry logging system, in passport offices and at airports to associate the visa details of a person upon arrival, and even to enrol the entire population of a country's legal residents and immigrants. The sophistication of the system, along with the very low false rejection rate, has made it reach the pinnacle of biometric security systems by being both reliable and secure.

## 2. IMAGE ACQUISITION

Before the process of enrolment via image acquisition beings, it is vital to understand how the eye works as a biological organ, as it helps in determining the specifications of the sensor used to capture the image itself.Furthermore, iris recognition is not synonymous with retina scanning, and while both are ocular-based, the former makes use of a high definition photographic detail of the person's iris which is used to examine its unique structure. Retina scanning on the other hand is potentially hazardous as the blood vessels are exposed during examination under relatively high intensity light.

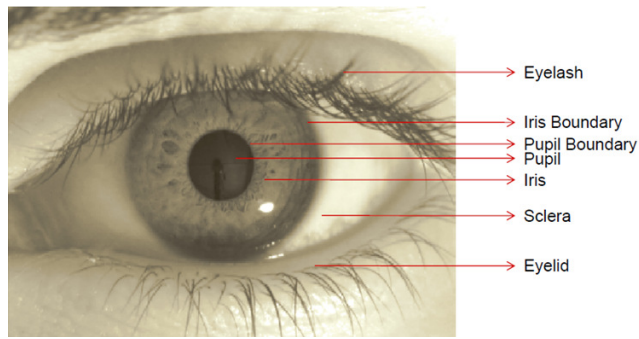

Figure 1. The anatomy of the human Eye, highlighting the Iris and the pupil

The Iris, as discussed in the previous section, is consistent of an extremely feature-rich texture – the trabecular meshwork. The part itself is enveloped by the inner pupillary boundary and the outer limbic boundary around the Iris. This changes in shape with the help of internal muscles in the eye to regulate the contraction and expansion of the pupil under different ambient conditions. The colour of the iris is a result of the melanin content, hence is different shades depending upon its concentration. This colour is not a part of the features to be extracted for examinations, rather the melanin interferes with the image if taken by optical sensors using wavelength of visible light. The melanin content peaks in the UV band in the absorption spectrum at 350nm, and with visible being at 400-700nm [3], the dark eyes (hazel, with high density of melanin) in those bands look despondent because of the low albedo and iris images dominated by the specular corneal reflection (mirror-like) cannot be processed very well. The blue colour pigmentation is naturally a resultant of the long wavelength light penetrating the anterior layer and the stroma. But the far end of the spectrum contains the NIR (Near Infra-Red) region, which has a low absorbance coefficient, hence making pigmentation/melanin density irrelevant. Hence most of the commercial iris recognition camera sensors are NIR, unlike the IR imaging used by retina scanners.

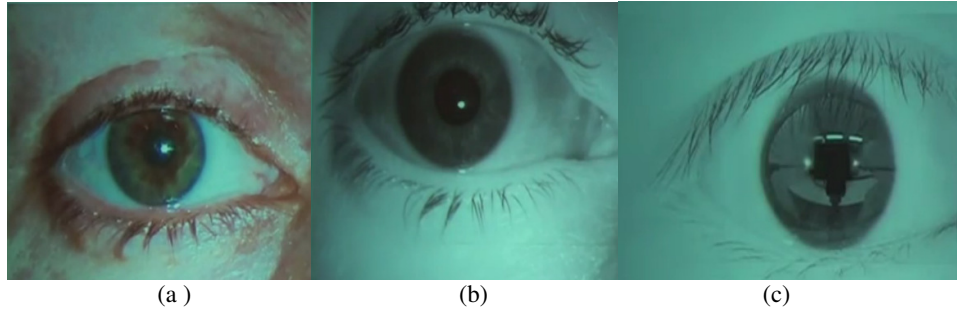

(a )                                    (b)                                    (c)

Figure 2.from left to right (a) photograph of the human eye taken under visible light in constrained/indoor conditions (b) the same eye captured with NIR camera (c) the evidence of corneal reflections, with image taken under visible light in un-constrained environments.

The basic trick in computer vision or imaging systems is to distinguish the difference between Lambertain iris and specular corneal components of reflections. The curved cornea surface obeys the Snell's Law and absorb the ambient light having a broad band of wavelengths. The NIR imaging camera has a blocking filter which cuts off the shorter wavelengths in the remaining band of light allowed to pass through, scrubbing out the ambient reflections. The image acquired by the NIR camera, used for unobtrusive imaging at a distance of say 1metre, is hence much more descriptive of the actual iris image, as even darkly pigmented irises reveal rich and complex features.

## 3. LITERATURE REVIEW

The history and the significant progress on iris biometric systems dates back to approximately 2001. But the idea itself is over a 100 years old, when the French artists Bertillion[4] mentioned the uniqueness of the features of an iris in his thesis "La couleur de l'iris. Revue scientifique". The automated algorithms were developed with the patent for an unimplemented design by Flom and Safir. They very elaborately designated the use of a highly monitored, occlusion-free environment, using a headrest and a manually operator for a fixed gaze at the subject's eye[5].
The illumination/ ambient properties were changed to determine the change in radius and size of the pupil as it contracted and expanded. They were amongst the first to suggest the various operators and tools in image analysis like corner detection, circular Hough transform and ways to extract the information from the iris. They even mentioned a given threshold for the intensity values to vary within, which was constant for all the images in the database.

The most notable pioneers in Iris Recognition systems with progressive and revolutionary algorithms are- Dr. John Daugman for first patenting a design using the integro-differential operator for Iris boundary localization and the rubber-sheet model for normalization and Wildes et al. developed an algorithm using circular edge maps to compute the boundary and a Hough transform to detect circles.

### 3.1. Daugman's Algorithm

In 1994, the most stable work on an iris biometric recognition system was evolved from the patent and publications by Dr. John Daugman [1] who described the functionality of this system in acute detail.

The biometric system also evolved with respect to the numerous operators used in the algorithm. Similar to the work done on face recognition systems and the speed cameras seen on the streets, the eyes were searched using a "deformed template". The primary part of iris localization included the segmentation and clear definition of the pupillary (inner) and the limbic (outer) boundaries. These were defined using a definite operator known as the integro-differential operator, which searched for boundaries in a given parameter space[1], [6].

$$max_{(r,x_0,y_0)} \left| G_\sigma(r) * \frac{\partial}{\partial r} \oint \frac{I(x,y)}{2\pi r} ds \right|$$

Where $G_\sigma(r)$ is thesmoothing function and I(x, y) the image in terms of the representative coefficients of the intensity values of the circular bound region within the x,y parameter space, with $x_0, y_0$, r being the circular and radial coordinates in the plane. These coordinates are maxed out within the measurement of the pupillary and limbic boundaries defined by the iris and pupil contour, but with the assumption that the potential illumination of the pupil is the maximum gradient circle (practically measured to be 0.8 minimum).

This formula was also experimented with and evolved into a much more sophisticated design, with the inclusion of images with the iris having an off x-axis gaze being a permitted entry[7]. Daugman's algorithm used the rubber sheet model, a method of mapping the external polar coordinates on a circular plane to transform it into a rectangular extracted iris region, irrespective of the noise factors like the eyelids and the eyelashes, which were excluded later.

For feature extraction, he used the 2D Wavelet operators to disintegrate the given image and re-assemble it by marginally reducing the size of the image, without consistently reducing the amount of image stored. After texture analysis, the information is matched using the Hamming distance, which is essentially a difference between the two iris code segments, with the word 'iris code' being coined by Daugman himself as a representation of the iris texture in binary stamp format. He is acclaimed as the father of Iris biometric systems.

## 3.2. Wilde's Algorithm

Wilde et al. [2]is another prominent scientologist who headed the project at Sarnoff Labs to develop an iris biometric system, with a technical approach slightly distinct from Daugman's. In 1996 and 1998, he had two patents which constituted of a unique acquisition system as well as a slightly less consistent but very effective automated iris segmentation method.

Instead of using a NIR video camera to capture a digitalized image, they have used a standard high resolution camera but with a distinct diffused light source or also described as "a low light level camera".

The iris localization is another distinction. While Daugman used the integro-differential operator, Wildes uses a much more primary route of firstly calculating the binary edge map of the image and then using Hough transform and the relative accumulator function to calculate the intensity levels of pixels constituting a circle, or if the image is distorted by noise, arcs. This algorithm helps detect the pupillary and limbic boundary contours which are then segmented and used the segmented image is sent for feature extraction.

The second distinct method is the usage of a Laplacian of Gaussian filter over multiple overlapping stages in order to produce a template for feature extraction instead of using the compressional methods like Haar wavelet decomposition or 2D Gabor filters.The matching is done by computing the normalized correlation as a measure of the similarity and distinctiveness of two iris codes.

## 4. IMPLEMENTED METHODOLOGY

The algorithm as a whole is divided into four distinct steps leading up to the matching, using a hybrid approach implemented using the CASIA database V 3.0 Iris Interval and practically verified using MATLAB R2014aStudent version. The system developed consists of the following steps: (1) Image Pre-processing using histogram matching, thresholding and canny edge operator (2) Localization of pupillary and limbic boundary using Circular Hough Transform (3) Iris Normalization using Daugman's rubber sheet model (4) Feature Extraction using Haar wavelets and binary encoding.

### 4.1. Pre-processing Techniques

Due to the presence of ambient variations in distinct images with varying levels of illumination, the histogram of each individual image needs to be have at least two distinct peaks which represent the distinct greyness intensity level in the image, the darkest part inarguably being the pupil. After selecting a particular image with a bi-modal histogram, the other images are matched taking the former as reference. The matching will usually induce contrast enhancement and the image is blurred later using a Gaussian filter.

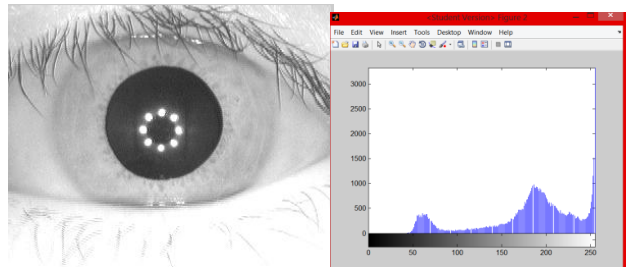

Figure 3. from left to right (a) The reference iris image (b) The histogram of the corresponding unequalized image with distinct peaks

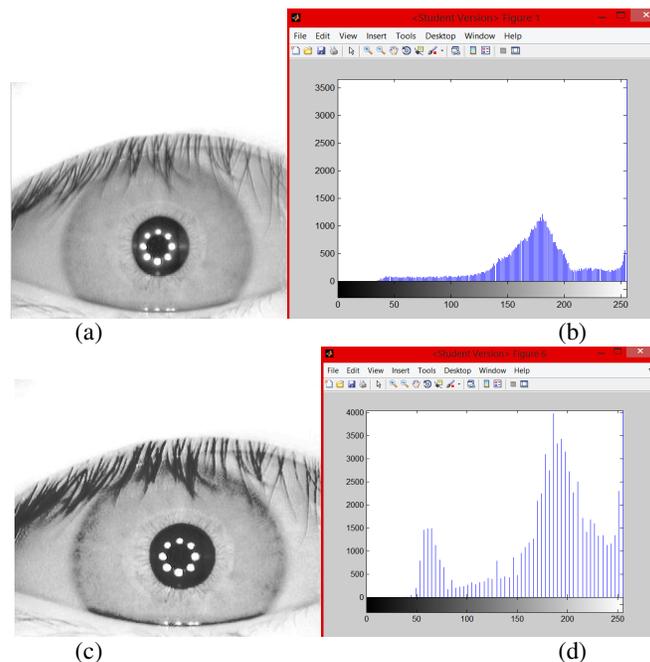

Figure 4. From top left to bottom right (a) The Iris image of the enrolled image (b) the corresponding unaltered histogram (c) The image after smoothing and histogram matching (d) the altered bi-modal histogram

To make sure that the pupillary boundary is properly segmented, a threshold is placed. The threshold is introduced before the canny edge operator is used in order to define only the high transition of intensity levels between the pupillary boundary and the iris [8], [9]. The Gaussian filter helps blur all other noises as well, like the eyelashes and eyelid boundaries.

## 4.2. Boundary Segmentation

After the image has been thresholded and the histogram equalized, the pupillary boundary becomes easy to localize. The localization is a result of the canny edge operator which establishes the coordinates of the boundary[10]. These coordinates are better defined using circular Hough Transform and the centres and radius of the pupil are noted and stored. Circular Hough Transform (CHT) is transforms a given subset of binary edge points present into an accumulated array of votes in the parameter space. For each edge point, votes are accumulated in an accumulator array for all parameter combinations. The array elements containing the peak number of votes indicate the presence of the shape.

These centric and radial coordinates aid in locating the outer iris (limbic) boundary as well, which is harder to localize as the intensity level of the transition values is not high enough. Since the pupillary and limbic boundaries are almost concentric, the later boundary can be approximated by building concentric circles and the intensity values of the pixels lying over the edge of the boundaries of these circles are calculated and summed using an accumulator array[11]. The difference between the summed values of each consecutive circle is noted and the boundary having maximum variation in intensity as compared the previous one is the limbic periphery. The eyelashes and eyelids, if they are covering the Iris region will not be excluded at this stage.

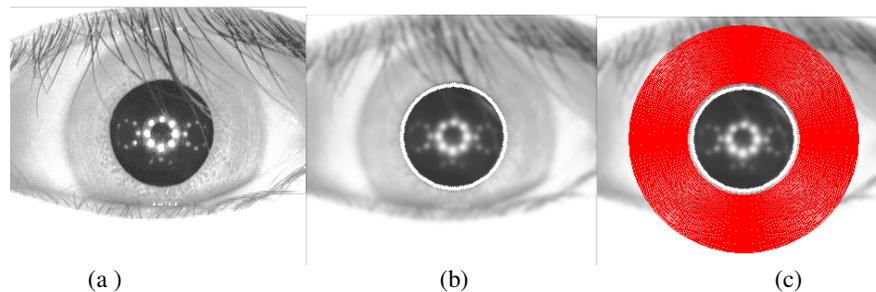

(a )                    (b)                    (c)

Figure 5. from left to right (a) resized iris image (b) image after pre-processing and pupil localization (white) (c) The concentric circles (red) produced by iterative solutions around the pre-detected pupil boundary

The next step is Normalization of the doughnut shaped regionby unwrapping the iris from its polar equivalent's Cartesian coordinates[6], [12].Though the word normalization is often mistaken to be synonymous with equalization or minimization of redundancy in statistical terms, in relations to this subject, normalization refers to the creation of a differently scaled/shifted version of the dataset. Hence, though the data remains the same, it is shifted by functions called pivotal quantities, whose sampling does not depend upon given parameters.The process of localization of an iris image has a major effect on the annular subset from the rest of the image, such that it is not linearized. Daugman had suggested a rubber sheet model, considering the surrounding and acquisition devise to cause the change in pupil's radius by dilation.

The Iris in unwrapped and all points along the edge contour map and within the boundary itself are converted into their polar Cartesian counterparts. The eyelashes and other noises are excluded by using a canny edge operator with a different threshold value.

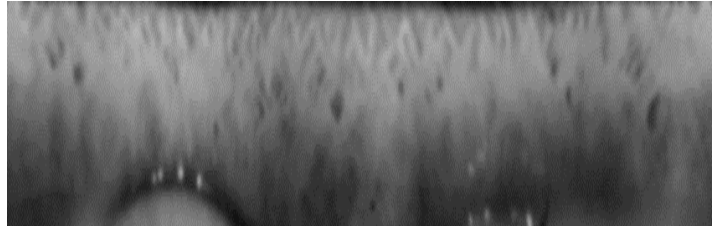

Figure 6. The normalized image obtained after contrast enhancement

### 4.3. Feature Extraction

The next process done on the normalized iris region is feature extraction [13]using Haar Wavelets[14].

Haar wavelet is a sequence of square-shaped functions which are rescaled as a part of some basis, usually a wavelet family. The sequence itself, very similar to the Fourier series, helps in the basic decomposition of a picture into its constituents, but without the use of sinusoidal functions. They essentially allow us to separate out the low frequencies and the high frequencies via an iterative method to allow for the compression of an image, also called as JPEG compression.

The basic idea behind the compression is to treat the given image (digital) as an array of numbers (matrix).Since the picture say, is a 256x256 pixel grey scale image, the image is stored in the form of a 256x256 matrix, with each matrix element being a whole number ranging from 0 (for black) to 225 (for white). The JPEG compression technique will allow us to keep on decomposing the image from 1x1 to 4x4 to 8x8 and so on till 256x256, assigning a given matrix to each black [15]. There will come an instance that the smaller elements of the decomposed array will have negligible value and hence can be neglected all together, allowing for compression of an image.

In mathematical 1D terms, Haar wavelets have an orthonormal basis for an interval, say, say, $L^2[0,1]$. When we multiply a unit function with a square function w (x) and integrate, the solution is zero. The compression happens as the time period of the function is squeezed. Eventually, we get smaller functions, removable with minimal loss of resolution [16].

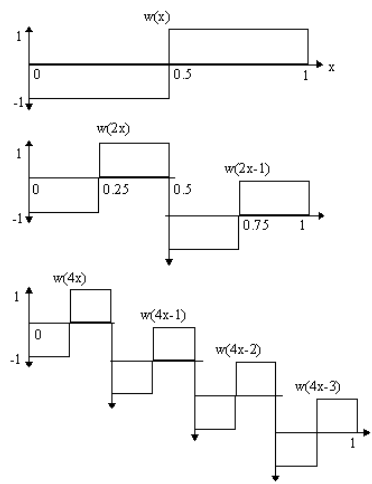

Figure 7. The 'squeezing' of an orthonormal function after iterative convolutions

In the figures above, it is clearly demonstrated how a signal, say w(x), which is a square function, when changed, or elongated using w(2x) will give us a similar shape function, but the functions are now squeezed when convolving the same signal to obtain w(2x-1), occupying the same space as the original signal, but compressed as the series progressed over lower frequency regions.

It is apparent that the iterative solutions obtained after constantly decreasing the frequency of the signals will leave us with smaller constituents again, which can be neglected. Though some data is hence lost in the compression technique, it is of very small value and can easily be taken back into consideration when comparing two strings and matching them under a given criteria.

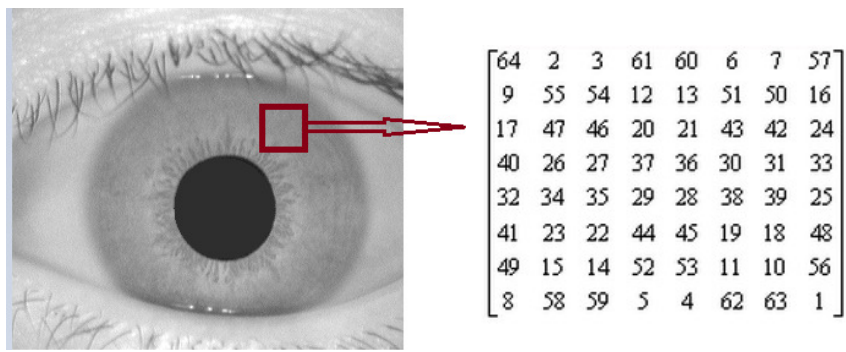

Figure 8. The the constituent elements in form a matrix for a decomposed 2D image

## 4.4. Binary Encoding

Haar wavelets allow for the compression of the image as well as helps us extract coefficients as approximation, vertical, horizontal and diagonal components. The former coefficients are decomposed by 3 levels and the results obtained as individual arrays from the column and row-wise summationcan be combined to form a singular matrix consisting of various negative and positive values.

To form the feature template, these values can be encoded in the form of binary numbers (1's and 0's), with the positive coefficients, including zero, are taken as 1, while the negative ones are 0. This gives a binary array which can be assigned to each iris image.

## 5. RESULTS AND DISCUSSION

The Iris biometric recognition systempresented in this paper was tested using the Iris images borrowed from the CASIA database who used an NIR homemade optical sensor to capture images with LED lighting. The procedure was verified using MATLAB R2014a on a 2.6 GHz processor with a dedicated graphics card. The time taken to localize the boundaries and encode the features of the normalized image into binary format was roughly 6.357 seconds per image at an average.

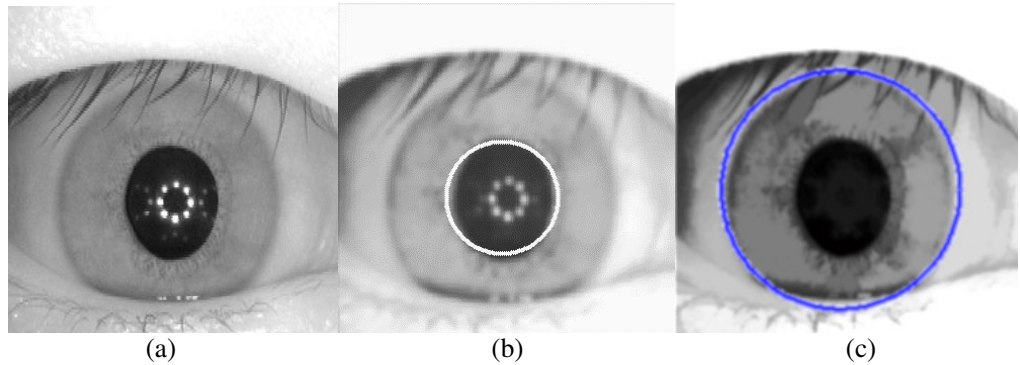

Figure 9. from left to right (a) An image with an erratic pupillary boundary (b) detection of pupillary boundary using CHT (white) (c) detection of limbic boundary (blue)

The primary results using the above algorithm has a success rate for recognizing and segmenting an iris region is 84% considering a definite but random subset of the database. The data obtained from this iris code was stored in separate folders in the database as the final output. This output is thus enrolled in the system and associated with the subjects' identity. If multiple iris codes from the same or different individuals are present they can be matched and authenticated using B-tree matching, but the algorithm presented in the paper is limited to enrollment and extraction of the Iris.

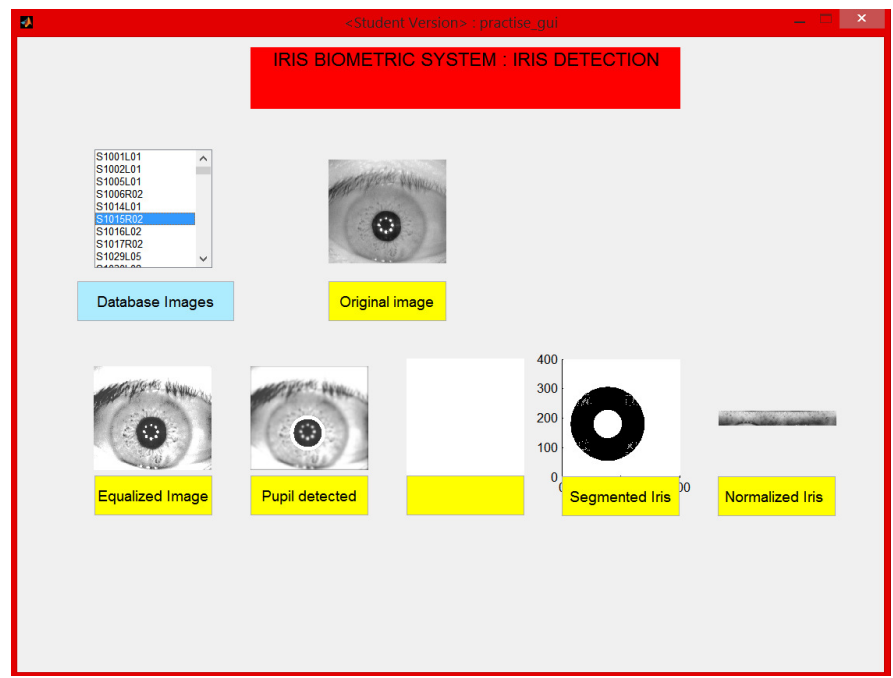

Figure 10. The GUI for the an independently developed Iris Biometric System showing the process till feature extraction

Figure 11. Some results showing the coordinates of boundaries and total elapsed time for
an image in the GUI

# 6. CONCLUSION

The above model and algorithm based system for creating a fairly competent Iris Biometric System was implemented using some basic image processing techniques like histogram equalization and Haar wavelets. The pre-processing methods were particularly important in removing the noise prematurely in order to make sure that they do not interfere with the identification of circular boundaries using Hough Transform. While the process itself just helps us to demarcate the Iris region, we need another method to convert the texture itself and encode it into computer-readable binary format- also called iris codes. The methodology, as proved by the results, is not susceptible to pupil dilation due to varying illumination, specular reflections, or erratic and inconsistent limbic or pupillary boundaries. This system can be validated with the help of B-tree matching with Hamming distance as a matching metric, which allows for comparison of two strings of iris codes. The Hamming distance can be varied with respect to results obtained after computing the rejection rate[17], [18] when comparing two analogous iris codes.

Iris based recognition systems are inarguably a very accurate and precise biometric technique to secure an individual's identity. While the system has itself been praised for being efficient and effective, scientists are still trying to improve the algorithm by publishing new research every year. This approach by using a hybrid method of pre-processing, boundary localization using CHT and feature extraction using Haar Wavelets is also another evolved notion of the same system currently being used all across the globe.

## ACKNOWLEDGEMENTS

I would like to express my deep and humble gratitude to Dr. Jagadish Nayak, my guru, who helped me implement this method efficaciously. Also I would like to thank the CASIA team for being kind enough to provide a huge database of Iris Images to the public.

# AUTHOR

Abhimanyu Sarin was born in New Delhi, India in 1992. He graduated from Birla Institute of Technology and Science, Pilani – Dubai Campus in 2014 with a B.Tech degree in Electrical and Electronics Engineering. His thesis coursework was completed during under graduation on Developing a fast and reliable Iris Recognition System for security purposes.

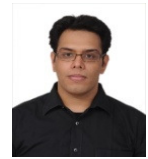